\documentclass[a4paper, 10pt, conference]{ieeeconf}      % Use this line for a4
\usepackage{FG2024}

\FGfinalcopy % *** Uncomment this line for the final submission

\IEEEoverridecommandlockouts                              % This command is only
\usepackage{amsmath}
\usepackage{amsfonts}
\usepackage{graphicx}
\usepackage{subcaption}
\usepackage{hyperref}
\usepackage{multirow}
\usepackage{authblk}

\def\FGPaperID{7} % *** Enter the FG2024 Paper ID here

\title{\LARGE \bf
Analyzing the Feature Extractor Networks for Face Image Synthesis
}

\usepackage{fancyhdr}
\begin{document}

\ifFGfinal
\thispagestyle{empty}
\author[1]{Erdi Sarıtaş}
\author[1,2]{Hazım Kemal Ekenel}
\affil[1]{\small Department of Computer Engineering, Istanbul Technical University, Istanbul, Türkiye}
\affil[2]{\small Department of Computer Science and Engineering, Qatar University, 
Doha, Qatar}
\pagestyle{empty}
\else
\pagestyle{plain}
\fi
\maketitle

\thispagestyle{fancy}

%%%%%%%%%%%%%%%%%%%%%%%%%%%%%%%%%%%%%%%%%%%%%%%%%%%%%%%%%%%%%%%%%%%%%%%%%%%%%%%%
\begin{abstract}

Advancements like Generative Adversarial Networks have attracted the attention of researchers toward face image synthesis to generate ever more realistic images. Thereby, the need for the evaluation criteria to assess the realism of the generated images has become apparent. While FID utilized with InceptionV3 is one of the primary choices for benchmarking, concerns about InceptionV3's limitations for face images have emerged. This study investigates the behavior of diverse feature extractors -- InceptionV3, CLIP, DINOv2, and ArcFace -- considering a variety of metrics -- FID, KID, Precision\&Recall. While the FFHQ dataset is used as the target domain, as the source domains, the CelebA-HQ dataset and the synthetic datasets generated using StyleGAN2 and Projected FastGAN are used. Experiments include deep-down analysis of the features: $L_2$ normalization, model attention during extraction, and domain distributions in the feature space. We aim to give valuable insights into the behavior of feature extractors for evaluating face image synthesis methodologies. The code is publicly available at \href{https://github.com/ThEnded32/AnalyzingFeatureExtractors}{https://github.com/ThEnded32/AnalyzingFeatureExtractors}.

\end{abstract}

%%%%%%%%%%%%%%%%%%%%%%%%%%%%%%%%%%%%%%%%%%%%%%%%%%%%%%%%%%%%%%%%%%%%%%%%%%%%%%%%
\section{INTRODUCTION}

Face image synthesis has become more popular since the introduction of Generative Adversarial Networks (GANs)~\cite{vanilla_gan}. This interest has further increased with the success of StyleGAN models \cite{ffhq,stylegan2}, which are capable of generating realistic and high-resolution images. However, despite these advancements, evaluating how realistic the generated images with such models remains a significant concern \cite{survey, roleoffid}. The core problem lies in the very definition, formulation, or quantification of being realistic. 

Human assessment may be helpful to evaluate small datasets \cite{neurosignal, hype}, yet become impracticable when the dataset size exceeds a few thousand. A more standardized approach calculates the closeness of the real (target) and the generated (source) image domains. This involves employing a feature extractor network to extract latent vector representation of images. Then, feature space representations are used to calculate the distance between the source and target domains \cite{inceptionscore, fid, kid, pNr}. 

Although the popularity of the evaluation metrics has differed in time, the InceptionV3 \cite{inception} model has been the de facto choice as a feature extractor after the introduction of the Inception-Score (IS) \cite{inceptionscore}. The researchers have focused mainly on improving calculations for evaluation metrics yet disregarded the feature extraction networks. Even though the CLIP \cite{clip} model has emerged recently \cite{survey, roleoffid}, the InceptionV3 network remains the dominating option. However, \cite{roleoffid} illustrated that InceptionV3 may attend to background parts outside the face in the images. This brings doubt about the trustworthiness of the FID results using the InceptionV3 model. Despite various networks being employed for feature extraction in \cite{roleoffid,extractoryeni}, they utilized solely FID.

In this study, we examine four networks from diverse domains as feature extractors: InceptionV3 \cite{inception}, CLIP  \cite{clip}, DINOv2  \cite{dinov2}, and ArcFace \cite{arcface}. Besides FID, we employ Kernel Inception Distance (KID) \cite{kid} and Precision\&Recall (P\&R) \cite{pNr} as evaluation metrics. For the experiments, we use two synthetic datasets -- generated by StyleGAN2 \cite{stylegan2}, and Projected FastGAN \cite{projectedgan} -- and a real dataset -- CelebA-HQ \cite{celeba} -- while FFHQ \cite{ffhq} is used as the target domain. The effect of projecting the features on the unit-sphere, $L_2$ normalization, is also investigated. Since the results are not directly comparable, we follow two assumptions to compare the behavior of those networks: \textit{1)} StyleGAN2 should surpass Projected FastGAN's performance \cite{roleoffid}; \textit{2)} metrics should give successful results when the source domain is also real. Moreover, we utilize visuals for deeper analysis regarding where the network attends during feature extraction and how these features are distributed in the feature space. 

The primary goal of this study is to provide an understanding of the feature extraction networks' impact on facial image synthesis evaluation. Even though this study specifically focuses on the face domain and utilizes GAN-based methods for generating the face images, the insights are still applicable to the other domains, e.g., bedrooms~\cite{otherdomain}, and tasks, e.g., domain transfer~\cite{pix2pix}, or for evaluating different synthesis methodologies \cite{vae,ddpm,idiff}.

%%%%%%%%%%%%%%%%%%%%%%%%%%%%%%%%%%%%%%%%%%%%%%%%%%%%%%%%%%%%%%%%%%%%%%%%%%%%%%%%
\section{RELATED WORKS}

The introduction of Generative Adversarial Networks \cite{vanilla_gan} revolutionized the field of image synthesis by generating highly realistic images. Even though no satisfactory results were obtained from this prior work \cite{vanilla_gan}, its methodology formed an essential framework for many tasks~\cite{survey}, such as conditional data generation \cite{conditional_gan} and domain transfer~\cite{pix2pix}. Enhancing super-resolution results~\cite{pdmsr} or generating datasets \cite{biasdataset} can exemplify its diverse application areas. StyleGAN \cite{ffhq} was one of the first networks that generated realistic images with a satisfactory resolution. The results of StyleGAN have motivated researchers to study GAN-based methods in a wider range, and thus, the necessity of benchmarking the realism of generated images has emerged.

One of the primary problems for benchmarking is the evaluation metric \cite{survey, roleoffid}. Without a universally agreed-upon definition of realism, it is problematic to determine whether the generated images are realistic. Some studies preferred to use human assessment \cite{neurosignal, hype}, but this approach could not be applicable as the size of the dataset increases. Moreover, the human assessment is not reproducible since the same assessors could not be available, or their decisions may change over time. Human assessment also brings further problems like personal biases or loss of focus during the experiments. Therefore, it is necessary to define algorithm-based evaluation metrics. For this purpose, \cite{inceptionscore, fid, kid, pNr} proposed approaches to calculate the closeness between real (target) and generated image (source) domains.

Inception-Score \cite{inceptionscore} -- one of the earliest methods -- incorporates extracting the class (softmax) probabilities of generated images using the InceptionV3 \cite{inception} model trained on the ImageNet \cite{imagenet} and employs KL-Divergence on those class probabilities. Fréchet Inception Distance (FID) \cite{fid} is introduced as a more advanced metric using real data statistics. Instead of class logits, latent vectors extracted from InceptionV3 are utilized. Then, following the Gaussian latent space assumption, a statistical metric is calculated using the mean and covariance matrices. Kernel Inception Distance \cite{kid} disregards the dependency on that assumption and uses maximum mean discrepancy. Finally, the Precision\&Recall \cite{pNr} metric uses the individual feature distances.

Despite remarkable efforts in the evaluation metric studies, the InceptionV3 model is still the primary choice as the feature extractor. There is limited work studying the effect of using different feature extractor networks~\cite{roleoffid,extractoryeni,extractorWPnR}. Moreover, they mainly consider only FID as the metric, except \cite{extractorWPnR} used P\&R as well. Recently, \cite{roleoffid} pointed out that using the InceptionV3 model may be problematic for assessing the generated face images' realism as its training dataset does not contain the 'human' class. 

%%%%%%%%%%%%%%%%%%%%%%%%%%%%%%%%%%%%%%%%%%%%%%%%%%%%%%%%%%%%%%%%%%%%%%%%%%%%%%%%

\section{Evaluation Metrics}
This work uses FID, KID, and P\&R as the evaluation metrics. The FID score is calculated as,

\begin{equation}
FID = ||\mu_S - \mu_T ||_2^2 + Tr( Cov_S + Cov_T - 2\sqrt{Cov_S \odot Cov_T} ) % \circ de olabilir
\label{eq-fid}
\end{equation}

\noindent where $Cov_S$ and $Cov_T$ are covariance matrices; $\mu_S$ and $\mu_T$ are mean vectors calculated from source features $f_S$ and target features $f_T$, respectively. $Tr(X)$ is the trace operation, $ || x ||_2^2$ is the $L_2$ norm, and $\odot$ is element-wise multiplication.

Using a similar notation, the KID score is calculated as,

\begin{equation}
    \begin{aligned}
    KID = \mathbb{E} [\frac{\sum((p(\hat{f}_S,\hat{f}_S)+p(\hat{f}_T,\hat{f}_T)) \odot (1-I))}{s(s-1)} \\
    -\frac{2\sum(p(\hat{f}_S,\hat{f}_T))}{s^2}]
    \end{aligned}
\label{eq-kid}
\end{equation}

\noindent where $\hat{f}_S$ and $\hat{f}_T$ are the random subsets with size $s$, and $I$ is $s\times s$ identity matrix. The $p(x,y)$ is the kernel function,

\begin{equation}
    p(x,y) =  (\frac{x y^T}{d}+ 1)^3,
\label{eq-kernel}
\end{equation}

\noindent used to calculate the maximum mean discrepancy, where $d$ represents the dimension.

Precision and recall are defined in (\ref{eq-pNr}) which benefit from the region function $r(x,f)$ given in (\ref{eq-manifold}). The region function checks whether the source feature $x$ lies in the K-Nearest-Neighbour region of the target feature $y$ which is the closest target feature to source feature $x$.

\begin{equation}
    \begin{aligned}
    Precision &=  \frac{\sum_{x \in f_S} r(x,f_T) }{m} \\
    Recall &=  \frac{\sum_{x \in f_T} r(x,f_S) }{n}
    \end{aligned}
\label{eq-pNr}
\end{equation}
\begin{equation}
    \begin{aligned}
    r(x,f) = \begin{cases}
    1, \text{ if } || x-y ||_2^2 <q || NN_k(y)-y ||_2^2 \\
    \text{\: \: for at least one } y \in f \\
    0, \text{ otherwise}
    \end{cases}
    \end{aligned}
\label{eq-manifold}
\end{equation}

\section{Feature Extractors}

While being a pioneer model and a popular choice, InceptionV3 may not be optimal for specific domains like faces. Beyond the commonly used InceptionV3, this work examines various types of models as feature extractors considering the architecture, training dataset, and tasks. By utilizing these diverse models, we aim to understand their behavior in detail.

\begin{itemize}
    \item \textbf{InceptionV3:} Due to its early adoption and being a well-known CNN architecture, InceptionV3 remains one of the most commonly used models to evaluate image synthesis methods.
    
    \item \textbf{CLIP:} CLIP \cite{clip} comprises Transformer-based text and image encoders. The goal of the CLIP training is mainly to match the encodings of text and image pairs from the LAION-400M dataset. The image encoder part, which is a Vision Transformer (ViT)  \cite{vit} model (ViT-B/32 variant), is used for this work.
        
    \item \textbf{DINOv2:} DINOv2 \cite{dinov2} builds upon DINO \cite{dino} which is a self-supervised training method. DINOv2 comprises modifications in the training scheme and introduces a massive training dataset, LVD-142M. Furthermore, a more recent version \cite{registers} is proposed where 'registers' are introduced to the ViT models to fix the artifact problem. The newer version used the ImageNet22k dataset \cite{imagenet} instead of LVD-142M. We employed ViT-B/14 with four register architecture.
        
    \item \textbf{ArcFace:} Since we focus on evaluating the realism of generated face images, we investigate how a face recognition model would perform. Therefore, we used the ArcFace \cite{arcface} model with ResNet100 \cite{resnet} backbone trained on the MS1MV3 \cite{ms1m} dataset.
    
\end{itemize}

\section{EXPERIMENTAL SETUP}

To generate synthetic image datasets for experiments, two StyleGAN2 \cite{stylegan2} models trained on the FFHQ \cite{ffhq} dataset are utilized. One is the original StyleGAN2, while the other is the Projected FastGAN \cite{projectedgan} (hereafter referred to as ProjectedGAN). As our first assumption, we will suppose that StyleGAN2 exhibits superior performance than the ProjectedGAN according to \cite{roleoffid}. They utilized an ImageNet pre-trained discriminator to indirectly guide the generator towards learning ImageNet statistics. However, despite the improvements in FID, the visual inspection shows the opposite regarding the realism of the images.

Additionally, we use the CelebA-HQ \cite{celeba} dataset, besides the generated ones. The motivation for this experiment is that, if the evaluation metrics are able to measure the realism, one would expect to see successful results in this setup since the source data is also real. This forms our second assumption: metrics should give successful results when the source domain is real.

\begin{figure}[!t]
    \centering
    \begin{tabular}{ccccc}
        \multicolumn{5}{c}{\small FFHQ}\\
        \includegraphics[width=0.1\linewidth]{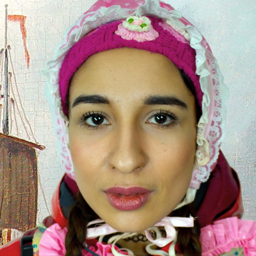}&
        \includegraphics[width=0.1\linewidth]{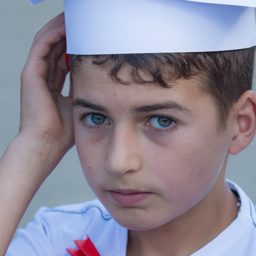}&
        \includegraphics[width=0.1\linewidth]{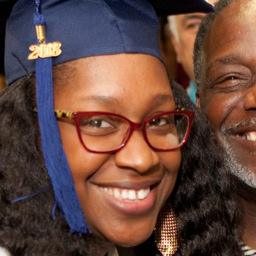}&
        \includegraphics[width=0.1\linewidth]{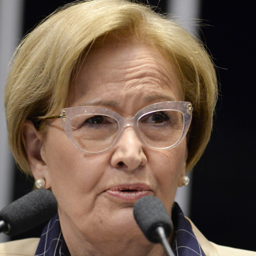}&
        \includegraphics[width=0.1\linewidth]{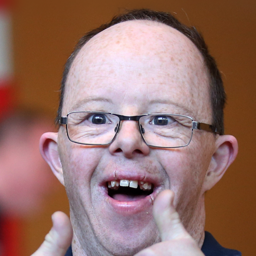}\\

        \multicolumn{5}{c}{\small CelebA-HQ}\\
        \includegraphics[width=0.1\linewidth]{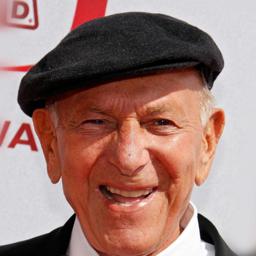}&
        \includegraphics[width=0.1\linewidth]{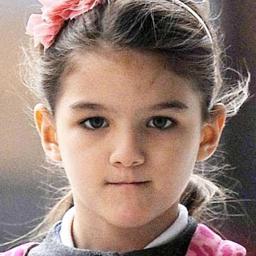}&
        \includegraphics[width=0.1\linewidth]{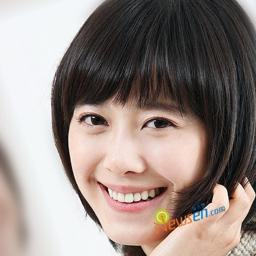}&
        \includegraphics[width=0.1\linewidth]{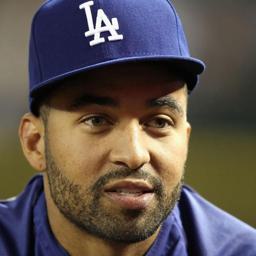}&
        \includegraphics[width=0.1\linewidth]{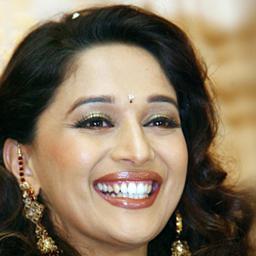}\\

        \multicolumn{5}{c}{\small StyleGAN2 }\\
        \includegraphics[width=0.1\linewidth]{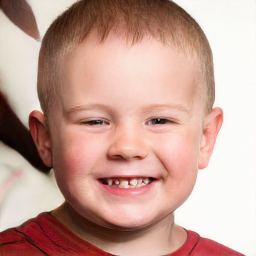}&
        \includegraphics[width=0.1\linewidth]{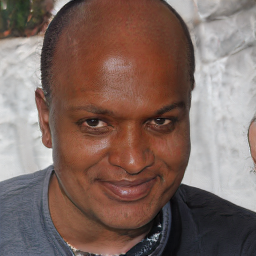}&
        \includegraphics[width=0.1\linewidth]{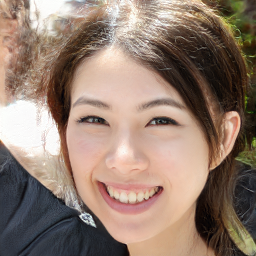}&
        \includegraphics[width=0.1\linewidth]{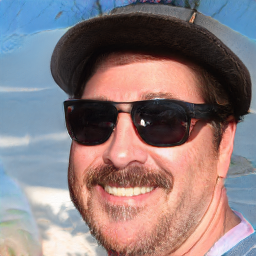}&
        \includegraphics[width=0.1\linewidth]{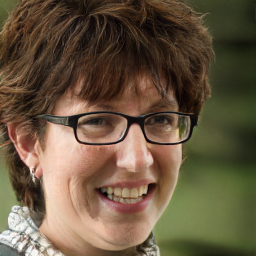}\\

        \multicolumn{5}{c}{\small ProjectedGAN }\\
        \includegraphics[width=0.1\linewidth]{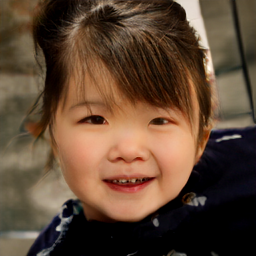}&
        \includegraphics[width=0.1\linewidth]{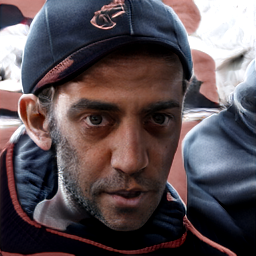}&
        \includegraphics[width=0.1\linewidth]{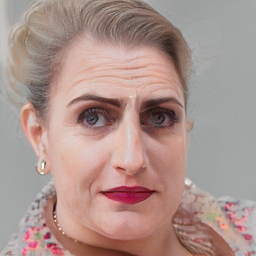}&
        \includegraphics[width=0.1\linewidth]{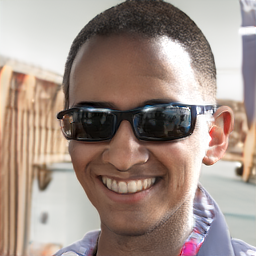}&
        \includegraphics[width=0.1\linewidth]{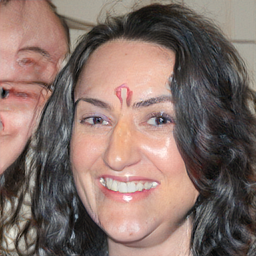}\\

    \end{tabular}
    \caption{\small Sample images from FFHQ\cite{ffhq}, CelebA-HQ~\cite{celeba}, and synthetic StyleGAN2 \cite{stylegan2}, ProjectedGAN \cite{projectedgan} generated datasets.}
    \label{fig:dataset_samples}
\end{figure}

The FFHQ dataset has 70k, and the CelebA-HQ has 30k $1024 \times 1024$ resolution face images. We downsample these to $256 \times 256$ resolution with bicubic interpolation. 50k $256 \times 256$ resolution images are generated for both generators. Sample face images are presented in Fig. \ref{fig:dataset_samples} -- see also the difference between StyleGAN2 and ProjectedGAN samples regarding their realism. The FFHQ dataset is our target, and the others are source datasets. Inspired by the usage in face recognition, $L_2$ normalization of the features is also investigated. The main results will be analyzed as 'original' and 'normalized'.

%%%%%%%%%%%%%%%%%%%%%%%%%%%%%%%%%%%%%%%%%%%%%%%%%%%%%%%%%%%%%%%%%%%%%%%%%%%%%%%%
\section{RESULTS}

\subsection{Comparison of StyleGAN2 \& ProjectedGAN}

According to the first assumption, we controlled whether any metric-feature extractor combination resulted in ProjectedGAN as the superior. Table \ref{tab-styleganXprojectedgan_results} presents the original features' results, whereas Table \ref{tab-styleganXprojectedgan_results_normalized} presents the normalized ones' results. 

\begin{table}[!t]
\caption{StyleGAN2 \& ProjectedGAN Evaluation Results}
    \centering
    %\hspace*{-0.8cm}
    \resizebox{0.93\linewidth}{!}{
    \begin{tabular}{|c||c||c||c||c||c|}
    \hline
        Feature     & Source & & & & \\
        Extractor   & Data   & FID $\downarrow$ & KID $\downarrow$ & P $\uparrow$ & R $\uparrow$ \\ \hline    
        \multirow{ 2}{*}{InceptionV3} & StyleGAN2 & \textbf{\textit{4.721}} & \textbf{\textit{0.0018}} & 0.684 & \textbf{\textit{0.413}} \\ \cline{2-6}
        ~           & ProjectedGAN & \textbf{\textit{4.520}} & \textbf{\textit{0.0009}} & 0.659 & \textbf{\textit{0.479}} \\ \hline
        \multirow{ 2}{*}{CLIP}        & StyleGAN2 & 2.173 & 0.0053 & 0.736 & 0.361 \\ \cline{2-6}
        ~           & ProjectedGAN & 4.111 & 0.0087 & 0.605 & 0.349 \\ \hline
        \multirow{ 2}{*}{DINOv2}      & StyleGAN2 & 53.307 & 0.1012 & 0.538 & \textbf{\textit{0.172}} \\ \cline{2-6}
        ~           & ProjectedGAN & 64.348 & 0.1298 & 0.481 & \textbf{\textit{0.180}} \\ \hline
        \multirow{ 2}{*}{ArcFace}     & StyleGAN2 & 0.825 & 0.0029 & 0.754 & 0.683 \\ \cline{2-6}
        ~           & ProjectedGAN & 2.820 & 0.0095 & 0.725 & 0.648 \\ \hline
        \multicolumn{6}{l}{ Results better with ProjectedGAN are marked by \textbf{\textit{bold and italic text}}.} \\
        \multicolumn{6}{l}{ $\downarrow$ indicates the lower score is better, and vice versa for $\uparrow$.} \\
        \multicolumn{6}{l}{ P represents the precision, and R represents the recall.} \\
    \end{tabular}}
\label{tab-styleganXprojectedgan_results}
\end{table}

\begin{table}[!t]
\caption{StyleGAN2 \& ProjectedGAN Evaluation Results - Normalized}
    \centering
    %\hspace*{-0.8cm}
    \resizebox{0.93\linewidth}{!}{
    \begin{tabular}{|c||c||c||c||c||c|}
    \hline
        Feature     & Source & & & & \\
        Extractor   & Data & FID* $\downarrow$ & KID* $\downarrow$ & P $\uparrow$ & R $\uparrow$ \\ \hline    
        \multirow{ 2}{*}{InceptionV3} & StyleGAN2 & 11.128 & \textbf{\textit{0.00352}} & 0.757 & \textbf{\textit{0.443}} \\ \cline{2-6}
        ~           & ProjectedGAN & 11.546 & \textbf{\textit{0.00193}} & 0.711 & \textbf{\textit{0.533}} \\ \hline
        \multirow{ 2}{*}{CLIP}        & StyleGAN2 & 19.645 & 0.03674 & 0.703 & 0.347 \\ \cline{2-6}
        ~           & ProjectedGAN & 36.468 & 0.05910 & 0.584 & 0.328 \\ \hline
        \multirow{ 2}{*}{DINOv2}      & StyleGAN2 & 91.171 & 0.11681 & 0.510 & \textbf{\textit{0.191}} \\ \cline{2-6}
        ~           & ProjectedGAN & 110.885 & 0.14928 & 0.428 & \textbf{\textit{0.215}} \\ \hline
        \multirow{ 2}{*}{ArcFace}     & StyleGAN2 & 1.089 & 0.00060 & 0.752 & 0.704 \\ \cline{2-6}
        ~           & ProjectedGAN & 5.739 & 0.00696 & 0.728 & 0.657 \\ \hline
        \multicolumn{6}{l}{ $^{\mathrm{*}}$ FID and KID scores are multiplied by 1000 for readability.} \\
        \multicolumn{6}{l}{ Results better with ProjectedGAN are marked by \textbf{\textit{bold and italic text}}.} \\
        \multicolumn{6}{l}{ $\downarrow$ indicates the lower score is better, and vice versa for $\uparrow$.} \\
        \multicolumn{6}{l}{ P represents the precision, and R represents the recall.} \\
    \end{tabular}}
\label{tab-styleganXprojectedgan_results_normalized}
\end{table}

Table \ref{tab-styleganXprojectedgan_results} shows the assumed relation for the InceptionV3 model with the FID metric. Furthermore, InceptionV3 is again the only model that prefers ProjectedGAN for KID as in FID. For the recall, only the CLIP and ArcFace models favor the StyleGAN2 generated images, yet no model favors ProjectedGAN with precision. In Table \ref{tab-styleganXprojectedgan_results_normalized}, the InceptionV3 model results better with StyleGAN2 when normalized features are used for the FID. If we look at the table column-wise, FID and precision did not prefer the ProjectedGAN for all the extractors. Moreover, CLIP and ArcFace did not prefer the ProjectedGAN for the row-wise inspection when using any metrics.

\subsection{Heat Map Analysis}

\begin{figure}[!b]
    \centering
    \begin{tabular}{cccc}
        \multicolumn{2}{c}{\small InceptionV3} & \multicolumn{2}{c}{\small CLIP}\\
        \includegraphics[width=0.1\linewidth]{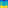}&
        \includegraphics[width=0.1\linewidth]{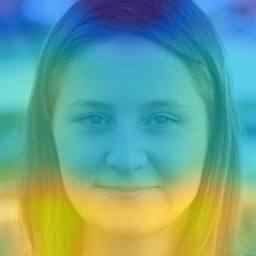}&
        \includegraphics[width=0.1\linewidth]{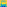}&
        \includegraphics[width=0.1\linewidth]{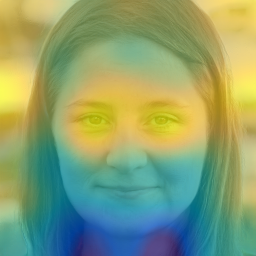}\\
        \multicolumn{2}{c}{\small DINOv2} & \multicolumn{2}{c}{\small ArcFace}\\
        \includegraphics[width=0.1\linewidth]{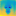}&
        \includegraphics[width=0.1\linewidth]{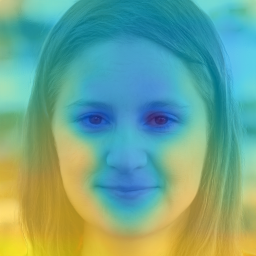}&
        \includegraphics[width=0.1\linewidth]{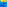}&
        \includegraphics[width=0.1\linewidth]{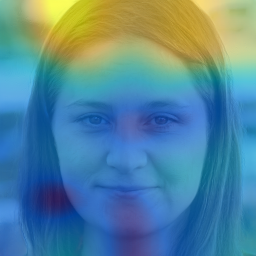}\\
        \small Low Attention & \multicolumn{2}{c}{\includegraphics[width=0.15\linewidth]{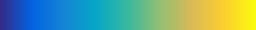}} & \small High Attention
    \end{tabular}
    \caption{\small Averaged heat maps using the generated images from StyleGAN2. The resolution of heat maps are 8x8, 7x7, 16x16, and 7x7 with the order of InceptionV3, CLIP, DINOv2, and ArcFace. In the heat maps, the left shows the original, and the right shows an overlay on a sample image. The used color map is given below, and attention increases from blue to yellow.}
    \label{fig:mean_heatmaps}
\end{figure}

Following \cite{roleoffid}, we extracted heat maps using the Grad-CAM \cite{gradcam} from StyleGAN2-generated images to analyze where the models attend. For the Grad-CAM algorithm, the layer before the last layer of the backbone was selected for the CLIP and the DINOv2, while the last layer was selected for the InceptionV3 and the ArcFace. The heat maps were averaged to analyze the expected attention area of the models on average. Fig. \ref{fig:mean_heatmaps} shows the aggregated heat maps. InceptionV3 attended to the lower part of the face where \cite{roleoffid} explains this by InceptionV3 misclassifying/founds objects like bow ties, which may be around the neck area, and focusing on them. Similarly, DINOv2 -- also trained on ImageNet -- attended the lower and outside face area. CLIP attended to the upper center of the face, which is the intended behavior. Whereas, ArcFace attended to the very top of the face. This may be because the upper face features are encoded into the top positions to separate from the lower face -- as cheating like in the artifact problem in \cite{dinov2}.

\subsection{Analysis with CelebA-HQ}

The evaluation metrics are designed to assess the success of generated images in terms of realism. This led us to the use of real data as the source domain. If the evaluating metrics are indeed able to measure realism, these metrics should produce successful results when we use a real dataset, i.e. the CelebA-HQ dataset. Tables \ref{tab-celeba_results} and \ref{tab-celeba_results_normalized} show the results. 

\begin{table}[!b]
\caption{CelabA-HQ Evaluation Results}
    \centering
    \resizebox{0.7\linewidth}{!}{
    \begin{tabular}{|c||c||c||c||c|}
    \hline
        Feature & & & &\\
        Extractor & FID $\downarrow$ & KID $\downarrow$ & P $\uparrow$ & R $\uparrow$ \\ \hline    
        InceptionV3 & 72.39 & 0.07311 & 0.194 & 0.130 \\ \hline
        CLIP & 34.08 & 0.14967 & 0.362 & 0.213 \\ \hline
        DINOv2 & 232.30 & 0.64587 & 0.473 & 0.461 \\ \hline 
        ArcFace & 22.77 & 0.40889 & 0.775 & 0.586 \\ \hline 
        \multicolumn{5}{l}{ $\downarrow$ indicates the lower score is better, and vice versa for $\uparrow$.} \\
        \multicolumn{5}{l}{ P represents the precision, and R represents the recall.} \\
    \end{tabular}}
\label{tab-celeba_results}
\end{table}

\begin{table}[!b]
\caption{CelabA-HQ Evaluation Results - Normalized}
    \centering
    \resizebox{0.7\linewidth}{!}{
    \begin{tabular}{|c||c||c||c||c|}
    \hline
        Feature &  &  & &\\
        Extractor & FID* $\downarrow$& KID* $\downarrow$ & P $\uparrow$ & R $\uparrow$ \\ \hline     
        InceptionV3 & 184.38 & 0.15625 & 0.198 & 0.170 \\ \hline
        CLIP & 321.05 & 1.13563 & 0.329 & 0.249 \\ \hline 
        DINOv2 & 386.71 & 0.66408 & 0.490 & 0.474 \\ \hline
        ArcFace & 26.27 & 0.07065 & 0.813 & 0.535 \\ \hline
        \multicolumn{5}{l}{ $^{\mathrm{*}}$ FID and KID scores are multiplied by 1000 for readability.} \\
        \multicolumn{5}{l}{ $\downarrow$ indicates the lower score is better, and vice versa for $\uparrow$.} \\
        \multicolumn{5}{l}{ P represents the precision, and R represents the recall.} \\
    \end{tabular}}
\label{tab-celeba_results_normalized}
\end{table}

For this experiment, we expect low scores for FID and KID and high scores for precision and recall. For Table~\ref{tab-celeba_results}, ArcFace gave the best scores except for KID, where InceptionV3 has the best KID score. For the FID and KID, DINOv2 gave the highest scores. Moreover, for Precision and Recall, InceptionV3 gave the lowest scores. When we move to normalized feature scores in Table \ref{tab-celeba_results_normalized}, ArcFace gave the best results for all metrics, suppressing others with margin. Furthermore, models differ for each metric when examining the worst results. For FID, KID, and Precision \& Recall (together), DINOv2, CLIP, and InceptionV3 give the worst scores, respectively. However, when we take a general look at Tables \ref{tab-celeba_results} and \ref{tab-celeba_results_normalized}, the results are much worse than both StyleGAN2 and ProjectedGAN, even though the data is real. This could be due to the fact that these metrics give scores on the closeness between extracted feature distributions of the source and target datasets, rather than the realism of the generated images. Therefore, the network and/or metrics used may be unsuitable for assessing realism.

\subsection{PaCMAP Analysis}

\begin{figure}[]
    \centering
    \begin{tabular}{cccc}
        \small InceptionV3 & \small CLIP & \small DINOv2 & \small ArcFace \\
         \includegraphics[width=0.125\linewidth]{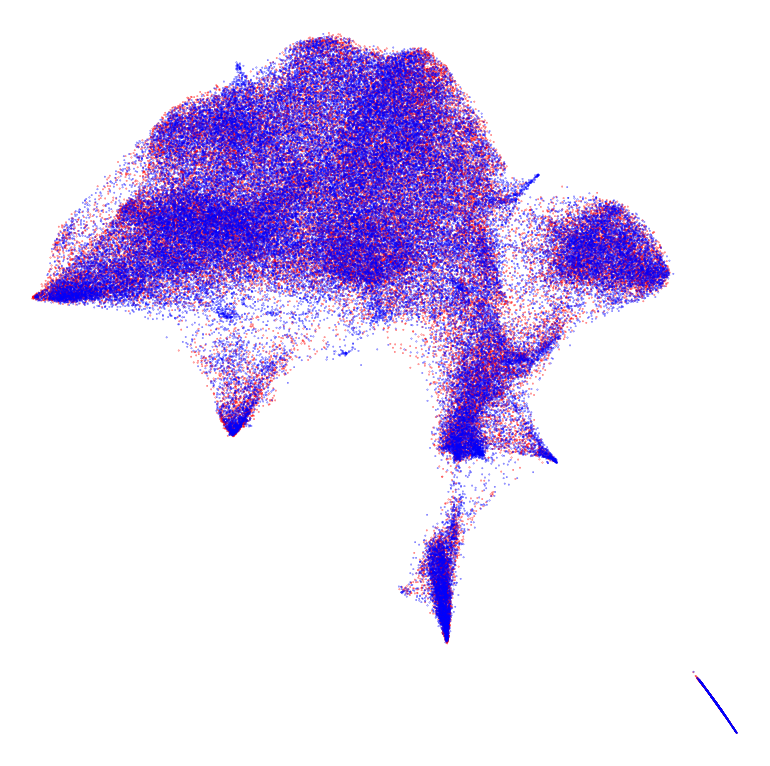} & \includegraphics[width=0.125\linewidth]{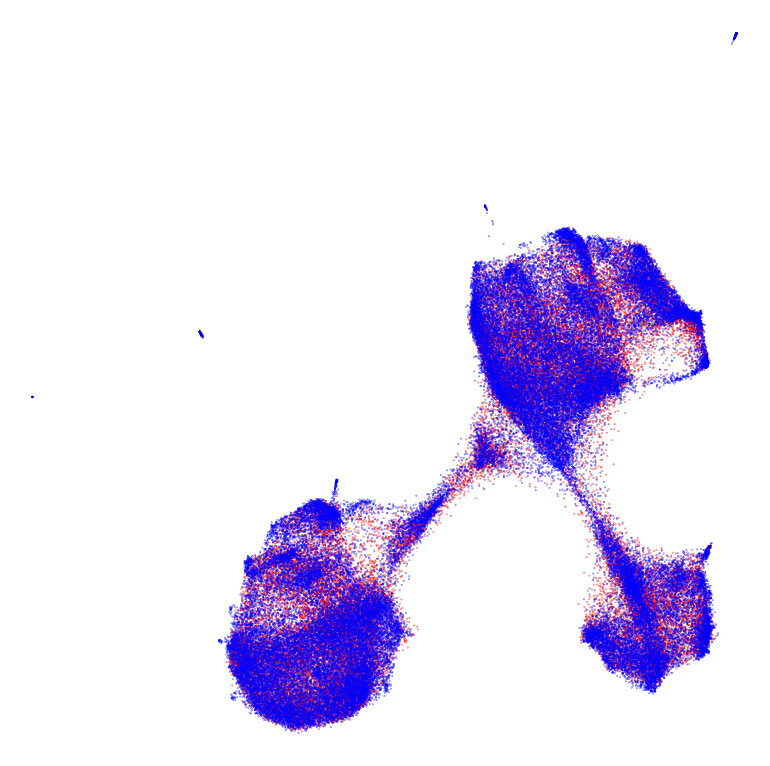}         &\includegraphics[width=0.125\linewidth]{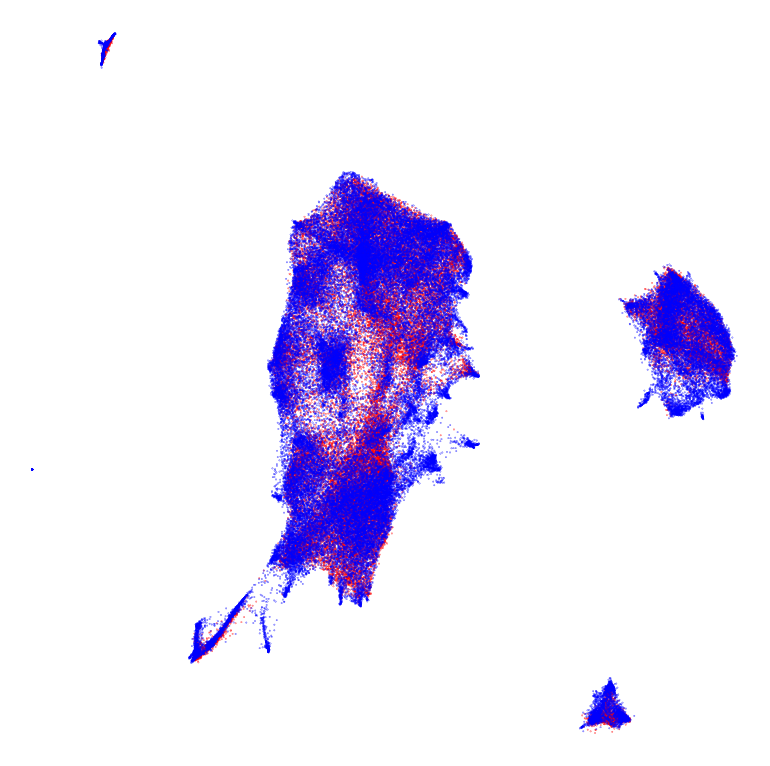} & \includegraphics[width=0.125\linewidth]{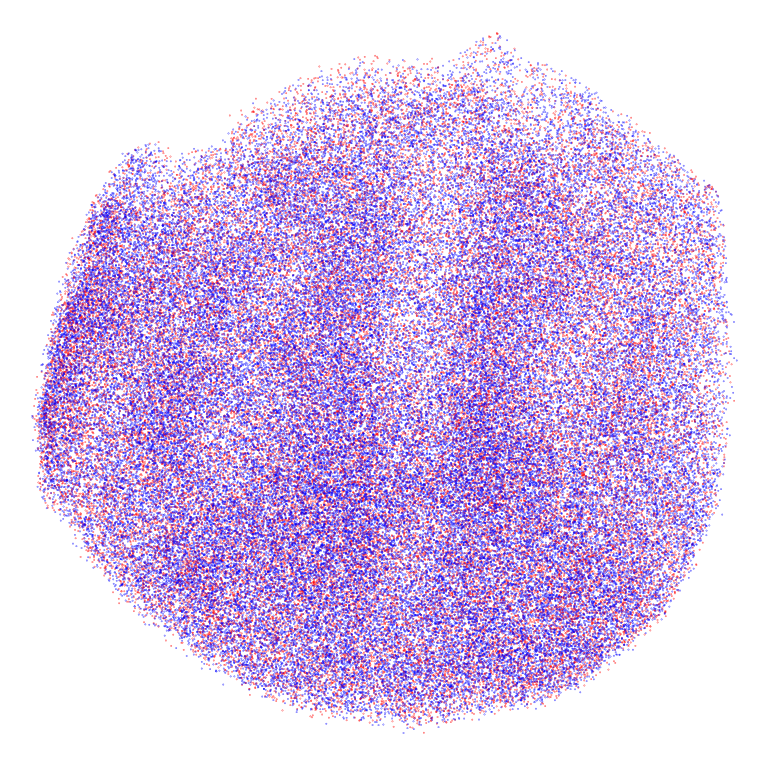}
        \\ \multicolumn{4}{c}{\small StyleGAN2} \\
         \includegraphics[width=0.125\linewidth]{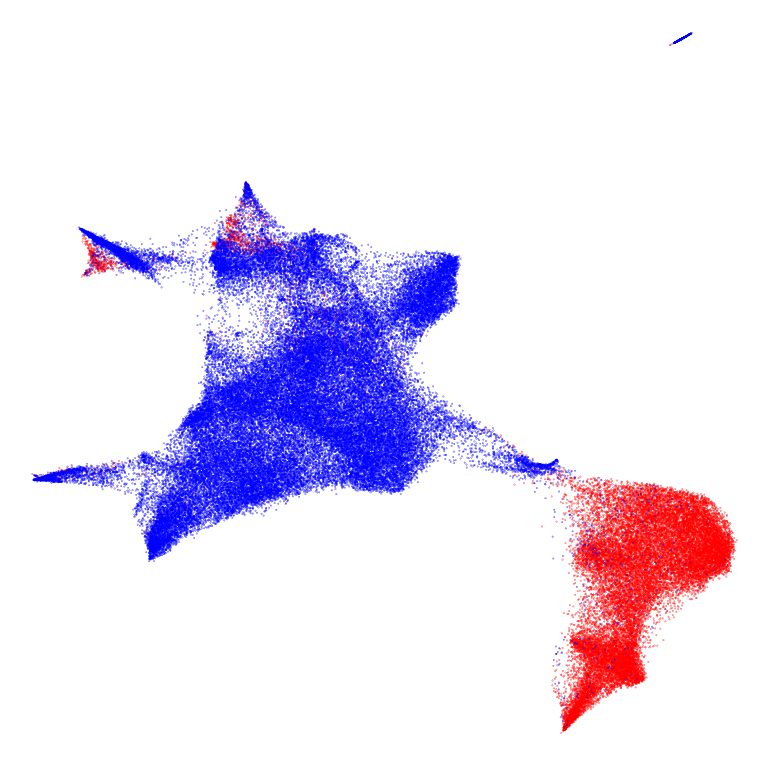} & \includegraphics[width=0.125\linewidth]{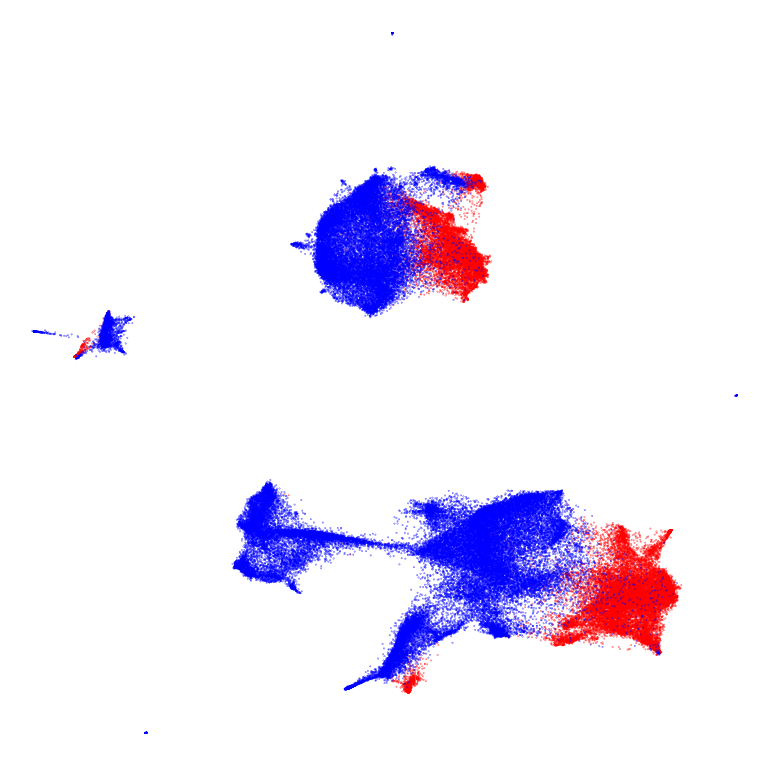}         &\includegraphics[width=0.125\linewidth]{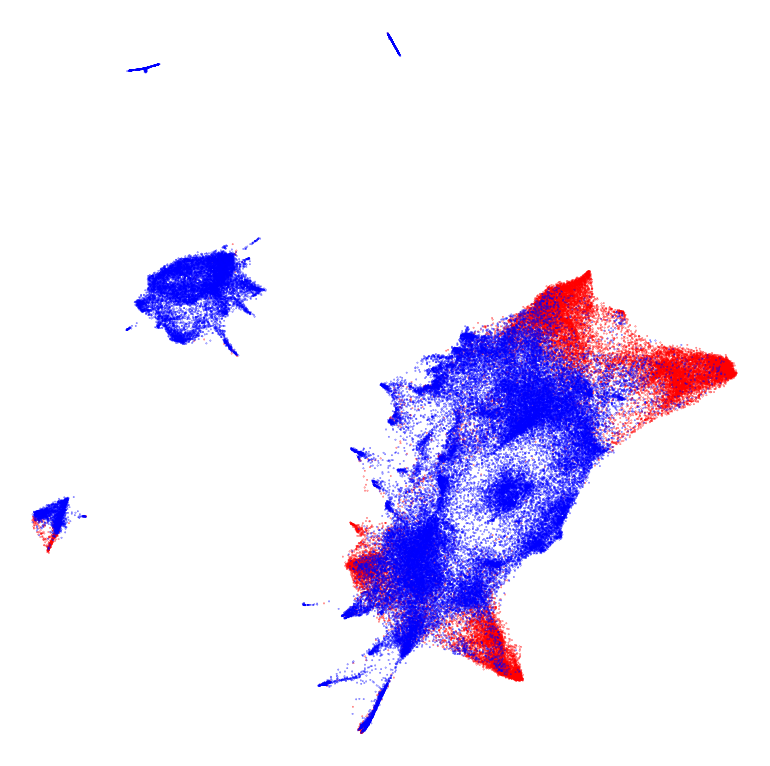} & \includegraphics[width=0.125\linewidth]{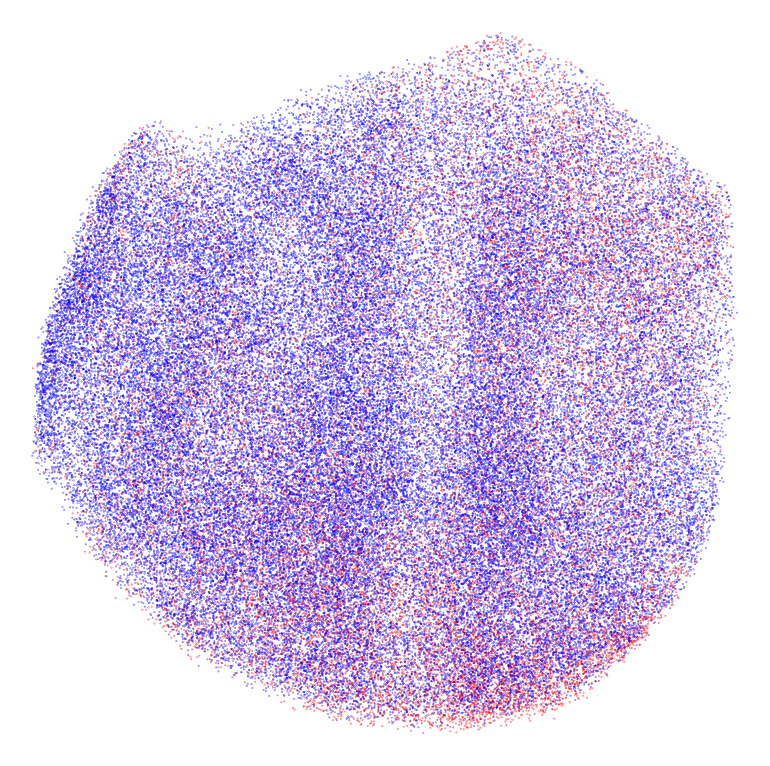}
        \\ \multicolumn{4}{c}{\small CelebA-HQ} \\
    \end{tabular}
    \caption{\small PaCMAP visualization results. In each sub-figure, blue dots represent the target, and red dots represent the source data points.}
    \label{fig:pacmaps}
\end{figure}

Due to unexpected results with CelebA-HQ, we wanted to investigate the domain relations in latent space. Thus, we used the PaCMAP \cite{pacmap} to create the 2D visualization of embeddings. Fig. \ref{fig:pacmaps} displays the 2D mappings. From the results, source and target data points are almost inseparable when using the features of StyleGAN2-generated images. While using the features of the CelebA-HQ dataset, only the ArcFace can be considered joined; we can see red-dominated or blue-dominated areas for the remaining network visuals. This explains the difference in results between ArcFace and the other models in Tables \ref{tab-celeba_results} and \ref{tab-celeba_results_normalized}. Moreover, the ArcFace also has an evenly distributed non-clustered feature space.

%%%%%%%%%%%%%%%%%%%%%%%%%%%%%%%%%%%%%%%%%%%%%%%%%%%%%%%%%%%%%%%%%%%%%%%%%%%%%%%%
\section{CONCLUSIONS}

This study comprehensively examines diverse feature extractor networks with various evaluation metrics focusing on the realism assessment of face image synthesis. The experiments are conducted on two real datasets, FFHQ and CelebA-HQ, and two synthetic image datasets generated by StyleGAN2 and Projected FastGAN (ProjectedGAN). Our study analyzes the results of FID, KID, and Precision \& Recall metrics with InceptionV3, CLIP, DINOv2, and ArcFace models. Furthermore, the evaluation calculations consider the extracted features' original and the $L_2$ normalized versions. For a deeper understanding, we examine the visual results from heat maps and the 2D mapping of embedding distribution to explain the behavior of models. Two premises guided the analysis of the results. Firstly, as implied by the previous work \cite{roleoffid}, ProjectedGAN should perform poorer than StyleGAN2 due to the realism of the images. Secondly, evaluation metrics were assumed to favor more realistic image sets. Thus, the results should be successful when real datasets, such as CelebA-HQ, are used for the source domain. 

From the experimental results, we can draw several conclusions. $L_2$ normalization may affect the model/metric preferences on the generated sets. Models focus on different parts of the facial area. Moreover, the synthetic image domain shows more remarkable similarity to the target domain than another real dataset. From the results, InceptionV3 and DINOv2 displayed poorer performance than others regarding the assumptions. Moreover, while CLIP demonstrated stable behavior, ArcFace showed promising performance for numerical results and embedding space yet its attention maps were perplexing. In our future work, we plan to include diffusion-based methods \cite{ddpm,idiff}.

%%%%%%%%%%%%%%%%%%%%%%%%%%%%%%%%%%%%%%%%%%%%%%%%%%%%%%%%%%%%%%%%%%%%%%%%%%%%%%%%
\section{ACKNOWLEDGMENTS}

\noindent \small We would like to thank Dr. Yusuf Hüseyin Şahin and Ziya Ata Yazıcı for their valuable comments. This work was supported by the TUBITAK BIDEB 2210 Graduate Scholarship Program, ITU BAP (Project No: 44120), and Meetween Project that received funding from the European Union's Horizon Europe Research and Innovation Programme under Grant Agreement No. 101135798.

% \noindent \includegraphics[width=1\linewidth]{metween.png}

%%%%%%%%%%%%%%%%%%%%%%%%%%%%%%%%%%%%%%%%%%%%%%%%%%%%%%%%%%%%%%%%%%%%%%%%%%%%%%%%

{\small
\bibliographystyle{ieee}
\bibliography{egbib}
}

\end{document}